\documentclass[runningheads]{llncs}
\usepackage{graphicx}
\usepackage{xcolor}
\usepackage{amsmath}

\usepackage{hyperref}
\begin{document}
\title{Algebra Error Classification \\ with Large Language Models}
\author{Hunter McNichols, Mengxue Zhang, Andrew Lan}
\authorrunning{McNichols et al.}
\institute{University of Massachusetts Amherst \\ 
Contact Email: \email{wmcnichols@umass.edu}}
\maketitle              % typeset the header of the contribution
\begin{abstract}
Automated feedback as students answer open-ended math questions has significant potential in improving learning outcomes at large scale. A key part of automated feedback systems is an error classification component, which identifies student errors and enables appropriate, predefined feedback to be deployed. Most existing approaches to error classification use a rule-based method, which has limited capacity to generalize. Existing data-driven methods avoid these limitations but specifically require mathematical expressions in student responses to be parsed into syntax trees. This requirement is itself a limitation, since student responses are not always syntactically valid and cannot be converted into trees. In this work, we introduce a flexible method for error classification using pre-trained large language models. We demonstrate that our method can outperform existing methods in algebra error classification, and is able to classify a larger set of student responses. Additionally, we analyze common classification errors made by our method and discuss limitations of automated error classification.

\end{abstract}
\section{Introduction}

Quality math education, particularly at a young age, is of crucial importance for students growing into an increasingly technology-driven world. Intelligent tutoring systems (ITSs) have demonstrated their effectiveness in improving math learning outcomes \cite{aleven2016instruction,fancsali2014context,roschelle2016online}. Many ITSs have a component that provides automated feedback for students while they solve math questions. This component enables teachers to provide personalized feedback at scale, since they can write feedback once that applies to many students. Moreover, in some ITSs, this feedback can direct a student to the precise error in their problem solving process. This direct, timely feedback enables immediate learning and helps students refine their math skills at their own pace \cite{pane2014effectiveness}.

One crucial part of feedback generation is the task of error classification, i.e., detecting student errors and the class corresponding to the error type. Once detected, the error class then informs the feedback generator which feedback is appropriate to provide to the student. This is especially important for open-ended math questions, where students have to reason step-by-step and one error can lead a student away from the correct solution. %\mh{I moved this part up}
A related line of work is automated grading of open-ended math responses \cite{mlp,erickson,sami,baral2022enhancing,edm22}, which is closely related to the automated short answer grading (ASAG) and automated essay scoring (AES) tasks \cite{aes,erater,irtasag,wang2019meta}. However, student error classification has an important difference from automated scoring: the former operates at a finer-grained level, focusing on individual solution steps, while the latter operates at a higher level, focusing on one score for the entire response.

Traditionally, error classification systems require a domain expert to develop hand-crafted rules that outline the different patterns of errors possible in a question \cite{pittcmu,rule,HeerenCog}. As a result, these systems are resource-intensive to create and do not generalize to responses that are not foreseen by the system designers. Data-driven methods have the potential to generalize to a wider range of responses, but few have been explored and most methods only process a final solution and not intermediate steps. The leading data-driven method to intermediate error classification relies on a tree embedding method \cite{MathOp}. This method is highly restrictive since it requires mathematical expressions in student responses to be converted to syntax trees \cite{treeEmbed}. Student solutions (particularly incorrect ones) are  not always syntactically valid, which means they cannot be processed by this method. Additionally, this method only uses equations as input and cannot be easily extended to include other student response information, such as natural language text or ITS-recorded interactions. %Our method overcomes both these limitations through use of pre-trained Large Language Models (LLMs).

\subsubsection{Contributions}
In this paper, we propose a method that overcomes the above limitations by using pre-trained Large Language Models (LLMs) for algebra error classification. Our contributions are:

\begin{itemize}
    \item We outline a method towards using pre-trained LLMs for algebra error classification, which enables us to handle any student response, regardless of syntactic validity.
    \item We demonstrate the effectiveness of this method through experimental evaluation using an algebra problem-solving dataset. We compare various pre-trained LLMs to the existing data-driven baseline, and show that BERT (and no other LLM) outperforms it.
    \item We showcase the flexibility of our method in that we can easily add more information into our classification system, which results in further performance improvement. We show how one can augment our method by adding ITS-recorded interactions to the input, or through Domain Adapation (DA) that incorporates further domain-specific pre-training to the LLM.
    \item We perform qualitative analysis on the cause of classification errors in our method and discuss avenues of future work.
\end{itemize}

\section{Methodology}

In this section, we outline how we formulate the problem of algebra error classification. Then, we outline our method, which uses a LLM to classify errors.

\subsection{The Algebra Error Classification Task}
The setup of the algebra error classification problem in an open-ended math setting is as follows. We have a set of predefined error classes for a category of math questions $C=\{c_1 \dots c_m\}$. We consider a solution that a student provides to a math question as a series of steps $S=\{s_1, \dots ,s_n\}$ (for simplicity of exposition, question statement can be included in the first step, $s_1$). A step encompasses a resulting intermediate equation and  descriptions of the process by which a student comes to this result. This description includes, but is not limited to, the application of a theorem, transformation, or intermediate calculation. Assume that, during step $s_t$, a student makes an error. We aim to identify the type of error $c_i$ made by a student at $s_t$ given the step history $S_t=\{s_1...s_{t}\}$.

Concretely, we consider a question from an introduction to algebra class:
\vspace{1pt}
\begin{align*}
\text{Solve\ for}\ x:\ x + 4 = 8
\end{align*}

With a student's solution:
\vspace{1pt}
\begin{align*}
    \text{Step 1:}\ &\ x + 4 = 8\\
    \text{Step 2:}\ &\ x + 4 + 4 = 8 + 4\\
    \text{Step 3:}\ &\ x = 12
\end{align*}

In this case, an error occurs at $s_2$, where a student adds to both sides instead of subtracting. While this is a valid algebra, it indicates the student is not progressing towards the solution. The task is to identify the class $c_i$ which corresponds to a ``wrong operation'' error. As input, we are given $S_t=\{s_1, s_2\}$. A classification could also be done for the addition error at $s_3$ using $S_t=\{s_1, s_2, s_3\}$.

It is important to note that step information is not limited to the equation. It could include any additional information available about how the student arrived at that equation. For example, notes in the margin between equations or ITS-recorded interactions which indicate student intent. 

\subsection{Our Method}
Our method to solve the algebra error classification task is to fine-tune a pre-trained LLM, which we detail below. 
 
\subsubsection{Model Architecture} We experiment with two main types of LLMs: those pre-trained on the masked language modeling (MLM) objective, such as BERT \cite{bert}, and those pre-trained on next token prediction in an autoregressive way, such as (GPT-2)\cite{gpt2}. Each type requires a different setup to be used as an error classifier. 

For MLM-type LLMs, the output contains the vector embedding of a \textbf{[CLS]} token. This token represents the contextual information of the entire input text sequence. As such, these models are designed for sequence classification. To form our classification head, we connect a linear layer from the \textbf{[CLS]} token to a probability space, which is the size of our number of error classes. For autoregressive models, such as GPT-2 \cite{gpt2}, there is no \textbf{[CLS]} token. To form our classification head, we instead connect a linear layer from the final hidden state representation (i.e. representation after the entire input sequence is encoded) to a probability space, which is the size of our number of error classes.

As input to the LLM, we tokenize a string that represents the step history. In our experiments, we try different methods of representing step history and compare the results. We note that input formulation is the most flexible part of our architecture and one of the main advantages of this method. First, it does not constrain the student responses to be a valid syntax tree. For example, response of ``$x + 3 = \text{unicorn}$'' is still a valid string. Second, it can be easily adapted to include further information. In our experiments, we only represent information from two steps $s_{t-1}$ and $s_t$. This representation is used to provide fair comparison to the leading data-driven method, which only considers the current and prior equation trees. However, this method could be easily extended to include the entire sequence history $S_t$ or responses to prior questions.

\begin{figure}[tp]
\includegraphics[width=\textwidth]{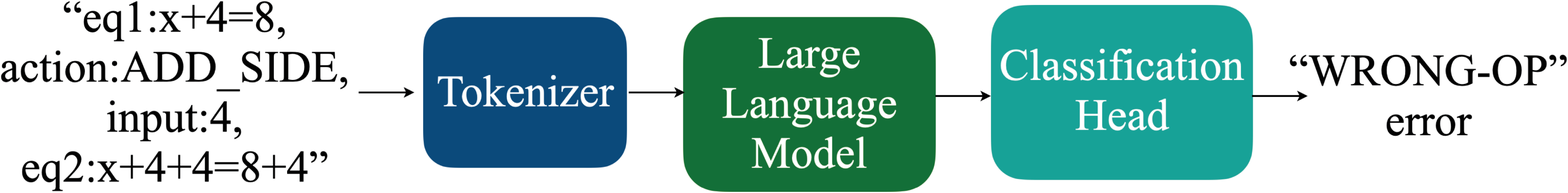}
\caption{Architecture diagram for our LLM-based method. Input is an arbitrary string for a set of steps. We depict the action information described in Section~\ref{sec:3.1} as additional step information. Output is a class corresponding to error detected in student response.} \label{arch}
\end{figure}

\subsubsection{Task-specific training}

To train our model for the task of algebra error classification, we minimize the standard cross-entropy loss\cite{MLPP} for error class predictions. We backpropagate this loss to learn the linear classification head from scratch and fine-tune the LLM. 
\label{AMPS}We also experiment with performing DA prior to fine-tuning. In DA, we perform additional pre-training on the LLM to adapt it to the domain-specific vocabulary of our task. This is performed with a domain-related dataset. In our case, we use a subset of the AMPS Mathematica dataset \cite{AMPS} containing solutions to PEMDAS questions generated by Mathematica. This DA step adds additional flexibility to our method, since we can readily adapt the LLM to more complex math questions, such as calculus or probability theory. This pre-training may make the LLM more familiar with advanced mathematical notation and reasoning. We leave this exploration for future work.

\section{Experimental Evaluation}

In this section, we detail our experiments and analyses. First, we show our quantitative experiments that compare the performance of our method to the performance of the baseline method. Then, we show quantitative experiments on two variations of our method, which use different strategies to include additional contextual information. Finally, we perform a qualitative error analysis on our best performing method.

\subsection{Dataset Details} \label{sec:3.1}

For our experiments, we use the CogTutor dataset accessed via PSLC Datashop \cite{koedinger2010data,CogTutor}. The dataset contains logs of students' step-by-step responses to questions in an Algebra I course. These logs were recorded during student interactions with an ITS named Cognitive Tutor. In Cognitive Tutor, students select from a set of predefined actions to manipulate an algebra equation. After selecting an action, the student inputs the expected resulting equation. Interactions are present in the dataset for each part of this process. We group the action interaction and the equation writing interactions into a single step.

The Dataset, in total, contains 130,823 interactions. These interactions span 9590 responses across 95 unique students. No demographic information is provided about the students. From the total interactions, we use a grouped subset of 5,744 steps that have a detected student error called \textit{BUG} \cite{MathOp}. In these cases, the ITS showed an automated feedback message to the student indicating an error made during the equation-solving process. These messages are based on a set of 92 predefined rules in the ITS. The feedback messages vary in length from short hints, such as ``Check your sign.'' to more descriptive feedback, such as ``You are dividing a positive by a negative. The result should be negative.''

\subsubsection{Baseline Labeling.} For our initial comparisons, we use the step labeling introduced in the leading data-driven method which has 24 distinct labels \cite{MathOp}. This labeling was present for 3,318 of the 5,744 \textit{BUG} steps; the remaining 2,426 steps contained equations which were not compatible with the Tree-Embedding approach utilized in the baseline method.

\subsubsection{New Labeling.} In subsequent experiments, where we compare variations of our method, we utilize our own labeling of the \textit{BUG} steps, which has 19 labels on 5,339 of the 5,744 \textit{BUG} steps. The remaining 403 steps were dropped from our analysis since they were either i) system errors not related to math operations or ii) rare errors that were present less than 30 times across the entire dataset. We make this new dataset, called \textit{CogTutorBugs}, publicly available \footnote{\href{https://github.com/umass-ml4ed/CogTutorBugs}{https://github.com/umass-ml4ed/CogTutorBugs}}. 

\subsection{Metrics and Baselines}
For error classification, our primary metric is classification accuracy, i.e., percentage of correctly predicted classes in the test set. We compare  
the following:

\sloppy
\begin{itemize}
  \item \textbf{TE+C} Best performing baseline method introduced in prior work \cite{MathOp}. This method converts intermediate equations in student response steps into tree-embeddings \cite{treeEmbed}, which are used to learn math operation embeddings. The operation embeddings are then used in a supervised learning manner for error classification.
  \item \textbf{GRU+C} Additional method explored in the baseline paper. Concatenates the equations together and passes the combined sequence character-by-character into a GRU \cite{gru}. Final output state is used for error classification. 
  \item \textbf{BERT} Bidirectional encoder representations from transformers (BERT) \cite{bert}. The predominant MLM-based LLM, with a format that is well-suited for classification tasks.
  \item \textbf{GPT-2} Generative Pre-trained Transformer 2 (GPT-2). An autoregressive LLM pre-trained for next sequence prediction instead of MLM. GPT-2 is well-suited for text generation tasks, but can be used for classification \cite{gpt2}.
  \item \textbf{MathBERT} A BERT-based LLM pre-trained on a corpus of mathematics textbooks and course material \cite{mathbert}.
  \item \textbf{XLM-RoBERTa} Multilingual variation of RoBERTa \cite{RoBERTa}, a BERT-based LLM pre-trained on specific natural language processing benchmark tasks such as question answering, reading comprehension, and natural language understanding. XLM-RoBERTa often outperforms BERT on these tasks \cite{xlmroberta}. 
\end{itemize}

\subsection{Implementation Details}

We perform five-fold cross-validation on the \textit{BUG} subsets of the CogTutor dataset. We use four folds for fine-tuning the LLMs and reserve the final fold for calculating test accuracy. For all experiments we train our models for 30 epochs. We report the mean and standard deviation in final-epoch test accuracy across the five folds.

All pre-trained LLMs are sourced from the HuggingFace \cite{huggingFace} transformers library. For fine-tuning we use the AdamW optimizer \cite{adamW}, batch size of 300, and learning rate of $5\cdot10^{-5}$. We do not perform hyperparameter tuning and use fixed hyperparameters across all models. All models were fine-tuned for 30 epochs. We found this number of epochs was sufficient to obtain optimal performance. All models were fine-tuned on a single NVIDIA RTX 8000 GPU. For each fold, fine-tuning took between 4 and 9 minutes, depending on the selected LLM.

\subsection{Results and Analysis}
\vspace{-10pt}
\begin{table}
\centering
\caption{Classification accuracy, using the error labeling in \cite{MathOp}, for our method with selected LLMs and the baseline methods. \textbf{Bold} indicates the best result.}\label{tab1}
\scalebox{0.9}{
\begin{tabular}{|l|l|}
\hline
{\bfseries Method} &  {\bfseries Accuracy $\uparrow$} \\
\hline
GRU+C &  $75.35\pm1.41 $  \\
TE+C &  $78.71\pm1.74 $  \\
BERT &  $\mathbf{80.68 \pm 1.1 }$  \\
GPT-2 &  $71.10\pm2.58 $  \\
MathBERT &  $72.66 \pm2.21 $  \\
XLM-RoBERTa &  $72.75\pm1.54 $  \\
\hline
\end{tabular}
}
\end{table}

\subsubsection{Comparison to Baseline Methods.} We report the accuracy of our method with each LLM and the baseline methods in Table~\ref{tab1}. We observe that BERT has the best performance across all LLMs and is a slight improvement over the leading baseline method (TE+C). Moreover, it achieves this performance without needing the tree-embeddings of the response equations used in TE+C. Perhaps surprisingly, other LLMs do not perform as well as either baseline method. This observation suggests that the choice of the LLM architecture and initial weights configuration is crucial for the success of LLMs in  classification tasks. BERT appears to be more robust in fine-tuning for the specific task of algebra error classification. On the contrary, GPT-2 isn't well-suited for classification likely due to its autoregressive design. XLM-RoBERTa and MathBERT are pre-trained on other tasks and show limited ability to generalize to our task. Their less competitive performance suggests that pre-training an LLM for one task may degrade performance in another, completely different task. We find it surprising that MathBERT, which is specifically designed for mathematical content, does not perform well on our task. One possible explanation is that MathBERT was pre-trained on textbook content, which can be different from problem-solving content. It also is pre-trained on student work in other ITSs, which may differ in structure to Cognitive Tutor.

\begin{table}
\vspace{-10pt}
\centering
\caption{Classification accuracy, using our new error labeling, for our method with selected LLMs. \textbf{Control} is the accuracy only with equation information in the input. \textbf{Action Inc.}\ is the accuracy with additional student action context provided by Cognitive Tutor ITS logs. \textbf{Bold} indicates the best result.}\label{tab2}
\scalebox{0.9}{
\begin{tabular}{|l|l|l|}
\hline
{\bfseries Language Model} &  {\bfseries Control $\uparrow$} & {\bfseries Action Inc. $\uparrow$} \\
\hline
BERT &  $82.02 \pm 1.09$  & $\mathbf{85.90 \pm 1.86}$ \\
GPT-2 &  $76.46\pm 1.47 $ & $74.21 \pm 2.08$ \\
MathBERT &  $76.08 \pm 2.05 $ & $81.27 \pm 0.32 $ \\
XLM-RoBERTa &  $77.73 \pm 2.59 $ & $84.23 \pm 2.65 $ \\
\hline
\end{tabular}
}
\vspace{-20pt}
\end{table}

\subsubsection{Use of Additional Context.} An advantage in using a LLM-based method for the error classification task is the flexibility of the input. Therefore, we experiment with including information about the student-selected action in the input (in addition to the step equations). This action information is recorded by Cognitive Tutor and provides context on student intent. We report the classification accuracy for the selected LLMs in Table~\ref{tab2}. For these experiments, we use our new feedback labeling. We observe that introducing action information improves the performance of all MLM models significantly. XLM-RoBERTa, in particular, has the largest improvement of 6.5\%. We note that GPT-2 actually shows a small performance deterioration when action information is included. This deterioration is small and could be explained by random noise in the experiments. Another potential explanation is that GPT-2 is pre-trained on natural language text, but the action information is programmatic, system log text. 

\begin{table}
\centering
\caption{Classification accuracy for BERT with various degrees of Domain Adaptation (DA). For example, BERT + 3 Epoch DA indicates 3 epochs of DA before classification fine-tuning and testing.} \label{tab3}
\scalebox{0.9}{
\begin{tabular}{|l|l|}
\hline
{\bfseries Model} &  {\bfseries Accuracy $\uparrow$} \\
\hline
BERT &  $82.02 \pm 1.09$  \\
BERT + 3 epoch DA &  $81.10 \pm 2.40$ \\
BERT + 10 epoch DA &  $81.00 \pm 1.72$ \\
\hline
\end{tabular}
}
\vspace{-5pt}
\end{table}

\begin{figure}[t]
\centering
\includegraphics[width=0.9\textwidth]{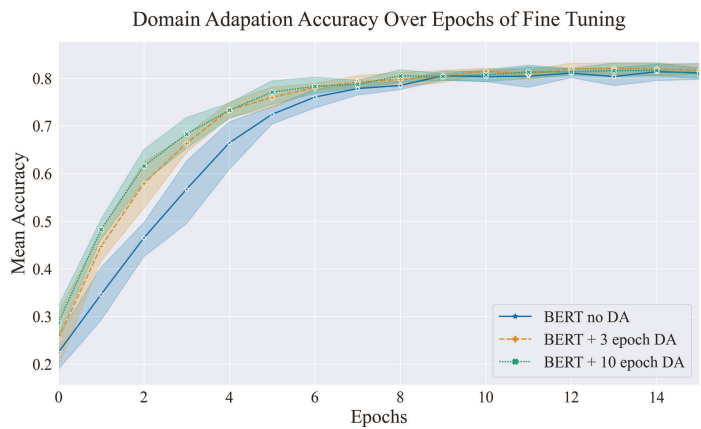}
\caption{Test accuracy comparison of DA methods. We observe only a slight improvement in initial epochs with DA and no significant difference in final performance (compared to no DA). The graph only shows first 15 epochs for clearer observation of early epochs. Shaded regions indicate range of accuracy across all folds. Best viewed in color.}\label{fig2}
\end{figure}

\subsubsection{Domain Adaptation} LLMs, such as BERT and GPT-2, are pre-trained on a variety of textual data but do not have domain-specific vocabulary for middle-school-level math. To adapt the LLM to better understand domain-specific vocabulary, we perform additional pre-training with in-domain but not task-related data. This step is done prior to task-specific fine-tuning and often called Domain Adaptation (DA). In Table~\ref{tab3}, we detail our experiments on performing DA with the PEMDAS subset of the AMPS Mathmatica dataset \cite{AMPS}. We observe no significant change in overall accuracy with the introduction of either 3 or 10 epochs of DA. One possible explanation is the small vocabulary size of our domain: only numbers and fundamental math operations: $+, -, /, *, =, ()$. However, we do see a non-trivial performance improvement in initial epochs of training as depicted in Figure~\ref{fig2}. This observation suggests that DA does provide some context on basic math vocabulary and mathematical reasoning, but this context is sufficiently learned from task fine-tuning within the first few epochs.

\begin{figure}[t]
\centering
\includegraphics[width=0.9\textwidth]{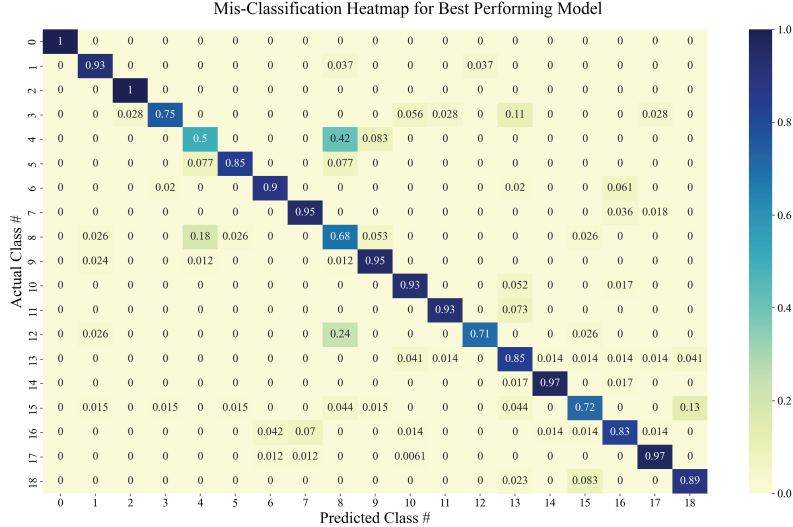}
\vspace{-20pt}
\caption{Mis-classification heatmap for qualitative error analysis of our method. Value in each cell is the ratio of the predicted class count to total number of actual class instances in the test set (each row sums to 1). Best viewed in color.} \label{fig3}
\vspace{-15pt}
\end{figure}

\begin{table}[t]
\centering
\caption{Example feedback templates for classes described in qualitative error analysis. The actual feedback students see have question-specific values for $A$, $B$, and $expression$.} \label{tab4}
\scalebox{0.90}{
\begin{tabular}{|c|c|p{7cm}|}
\hline
{\bfseries \#} & \bfseries{Class Label} & \bfseries{Sample Feedback} \\
\hline
4 & \text{REVERSED\_SIDES} & ``$expression$ is equal to $A$ minus $B$. You need to calculate $-A$ minus $B$.'' \\
8 & \text{WRONG\_OPERATION} &  ``$expression$ is equal to $A$ plus $B$. You need to calculate $A$ minus $B$.'' \\
12 & \text{FORGOT\_NEGATIVE} & ``You forgot the negative sign.'' \\
15 & \text{CHECK\_YOUR\_SIGN} & ``$expression$ Check your sign.'' \\
3 & \text{MULTIPLY\_TO\_SIMPLIFY} & ``Put the expression in its simplest form by performing multiplication on the right side.'' \\
13 & \text{SIMPLIFY\_FRACTION} & ``Simplify the fraction $expression$'' \\
\hline
\end{tabular}
}
\vspace{-10pt}
\end{table}

\subsubsection{Qualitative Error Analysis} We now analyze the cases when our best trained method (BERT + Action) mis-classifies student errors in a sample cross-validation fold. We show a mis-classification heatmap in Figure~\ref{fig3}. We observe two major causes of mis-classification: between two error classes where intent is ambiguous and between classes that are equivalent, but vary in level of specificity. Detailed information about each class mentioned is provided in Table~\ref{tab4}. 

\sloppy
The most frequent mis-classification is between error class \#4 and \#8 (REVERSED\_SIDES and WRONG\_OPERATION respectively). Class \#4, 41\% of the time, is incorrectly classified as class \#8. 18\% of the time the reverse mis-classification occurs. We observe that, in some scenarios, a REVERSED\_SIDES error is equivalent to a WRONG\_OPERATION error. Consider the question $y + 5 = -2$ and a student response $y = -3$. The student could have attempted to subtract 5 from both sides, but dropped the negative on -2. Equivalently, the student could have attempted to subtract -5 from both sides, erroneously thinking subtraction of -5 would cancel the 5 on the left hand side and confusing the rules of subtracting negatives. Both error classes can be appropriate, depending on the student's thought process. However, the true student intent is unknown, so it's ambiguous which class is correct.

\sloppy
Other frequent mis-classifications occur between classes that are almost the same but different in level of specificity. For example, class \#12 (FORGOT\_NEGATIVE) is 23.6\% of the time mis-classified as class \#8 (WRONG\_OPERATION). In the case of addition and subtraction, a sign error can be explained by the selection of an incorrect operation. In that case, an operation error is a more specific error than the sign error. A similar type of mis-classification occurs 13.2\% of the time from class \#15 to class \#18, and 11\% of the time from class \#3 to class \#13.

Through this qualitative analysis, we observe the most frequent mis-classifications of our method are explainable. They often occur because of ambiguous or overlapping classes. These observations suggest our method may be performing better than the accuracy metrics indicate. The observations also suggest that there is an upper limit to the performance possible on our chosen dataset, depending on how error classes are defined. Furthermore, the feedback selected by the rules of the Cognitive Tutor ITS should not be seen as the absolute truth. Without a perfect knowledge of student intent for each step of a response, it is impossible to have certainty about the true cause of student error. This is a general limitation which underlies all error classification systems.

\section{Discussion, Conclusion, and Future Work}

In this paper, we detailed a large language model-based method for the task of math error classification in open-ended questions. We now discuss our key observations and outline avenues for future work. First, we observe that our method, when combined with BERT, performs better than the best-performing, data-driven method for math error classification. Additionally, since it does not rely on syntactically correct responses, the method can operate on a wider range of student responses than the baseline method. Second, we observe that that incorporating additional information about student intent is easily achievable with our method and can provide a significant performance improvement. We hypothesize that providing further contextual information, such as student knowledge levels, will continue to improve performance, e.g., using models that can understand open-ended responses \cite{okt,mathgpt}. Third, we observe, through our qualitative analysis, that the errors made by our method are mainly due to ambiguous error class labels. This observation suggests an avenue for future exploration, where we extend our method to generate feedback, rather than classify errors.

\section{Acknowledgements}
The authors thank the NSF (grants 1917713, 2118706, 2202506, 2215193) for partially supporting this work.

\bibliographystyle{splncs04}
\bibliography{paper}

\begin{thebibliography}{10}
\providecommand{\url}[1]{\texttt{#1}}
\providecommand{\urlprefix}{URL }
\providecommand{\doi}[1]{https://doi.org/#1}

\bibitem{aleven2016instruction}
Aleven, V., McLaughlin, E.A., Glenn, R.A., Koedinger, K.R.: Instruction based
  on adaptive learning technologies. Handbook of research on learning and
  instruction pp. 522--560 (2016)

\bibitem{sami}
Baral, S., Botelho, A.F., Erickson, J.A., Benachamardi, P., Heffernan, N.T.:
  Improving automated scoring of student open responses in mathematics.
  International Educational Data Mining Society  (2021)

\bibitem{baral2022enhancing}
Baral, S., Seetharaman, K., Botelho, A.F., Wang, A., Heineman, G., Heffernan,
  N.T.: Enhancing auto-scoring of student open responses in the presence of
  mathematical terms and expressions. In: International Conference on
  Artificial Intelligence in Education. pp. 685--690. Springer (2022)

\bibitem{pittcmu}
Brusilovsky, P., Peylo, C.: Adaptive and intelligent web-based educational
  systems. International Journal of Artificial Intelligence in Education
  \textbf{13}(2-4),  159--172 (2003)

\bibitem{erater}
Burstein, J.: The e-rater{\textregistered} scoring engine: Automated essay
  scoring with natural language processing.  (2003)

\bibitem{gru}
Chung, J., Gulcehre, C., Cho, K., Bengio, Y.: Empirical evaluation of gated
  recurrent neural networks on sequence modeling. arXiv preprint
  arXiv:1412.3555  (2014)

\bibitem{xlmroberta}
Conneau, A., Khandelwal, K., Goyal, N., Chaudhary, V., Wenzek, G., Guzm{\'a}n,
  F., Grave, E., Ott, M., Zettlemoyer, L., Stoyanov, V.: Unsupervised
  cross-lingual representation learning at scale. arXiv preprint
  arXiv:1911.02116  (2019)

\bibitem{bert}
Devlin, J., Chang, M.W., Lee, K., Toutanova, K.: Bert: Pre-training of deep
  bidirectional transformers for language understanding (2018),
  \url{https://arxiv.org/abs/1810.04805}

\bibitem{erickson}
Erickson, J.A., Botelho, A.F., McAteer, S., Varatharaj, A., Heffernan, N.T.:
  The automated grading of student open responses in mathematics. In:
  International Conference on Learning Analytics \& Knowledge. p. 615–624
  (2020)

\bibitem{fancsali2014context}
Fancsali, S.E., Ritter, S.: Context personalization, preferences, and
  performance in an intelligent tutoring system for middle school mathematics.
  In: International conference on learning analytics and knowledge. pp. 73--77
  (2014)

\bibitem{HeerenCog}
Heeren, B., Jeuring, J., Sosnovsky, S.A., Drijvers, P., Boon, P., Tacoma, S.,
  Koops, J., Weinberger, A., Grugeon{-}Allys, B., Chenevotot{-}Quentin, F., van
  Wijk, J., van Walree, F.: Fine-grained cognitive assessment based on
  free-form input for math story problems. In: Pammer{-}Schindler, V.,
  P{\'{e}}rez{-}Sanagust{\'{\i}}n, M., Drachsler, H., Elferink, R., Scheffel,
  M. (eds.) Lifelong Technology-Enhanced Learning - 13th European Conference on
  Technology Enhanced Learning, {EC-TEL} 2018, Leeds, UK, September 3-5, 2018,
  Proceedings. Lecture Notes in Computer Science, vol. 11082, pp. 262--276.
  Springer (2018). \doi{10.1007/978-3-319-98572-5\_20},
  \url{https://doi.org/10.1007/978-3-319-98572-5\_20}

\bibitem{AMPS}
Hendrycks, D., Burns, C., Kadavath, S., Arora, A., Basart, S., Tang, E., Song,
  D., Steinhardt, J.: Measuring mathematical problem solving with the math
  dataset. NeurIPS  (2021)

\bibitem{rule}
Koedinger, K.R., Anderson, J.R., Hadley, W.H., Mark, M.A.: Intelligent tutoring
  goes to school in the big city. International Journal of Artificial
  Intelligence in Education  \textbf{8},  30--43 (1997)

\bibitem{koedinger2010data}
Koedinger, K.R., Baker, R.S., Cunningham, K., Skogsholm, A., Leber, B.,
  Stamper, J.: A data repository for the edm community: The pslc datashop.
  Handbook of educational data mining  \textbf{43},  43--56 (2010)

\bibitem{mlp}
Lan, A.S., Vats, D., Waters, A.E., Baraniuk, R.G.: Mathematical language
  processing: Automatic grading and feedback for open response mathematical
  questions. In: Proceedings of the ACM conference on learning@scale. pp.
  167--176 (2015)

\bibitem{okt}
Liu, N., Wang, Z., Baraniuk, R., Lan, A.: Open-ended knowledge tracing for
  computer science education. In: Conference on Empirical Methods in Natural
  Language Processing. pp. 3849--3862 (2022)

\bibitem{RoBERTa}
Liu, Y., Ott, M., Goyal, N., Du, J., Joshi, M., Chen, D., Levy, O., Lewis, M.,
  Zettlemoyer, L., Stoyanov, V.: Roberta: A robustly optimized bert pretraining
  approach (2019). \doi{10.48550/ARXIV.1907.11692},
  \url{https://arxiv.org/abs/1907.11692}

\bibitem{adamW}
Loshchilov, I., Hutter, F.: Decoupled weight decay regularization. arXiv
  preprint arXiv:1711.05101  (2017)

\bibitem{MLPP}
Murphy, K.P.: Machine learning: A probabilistic perspective. MIT Press (2021)

\bibitem{aes}
Page, E.B.: The imminence of grading essays by computer. The Phi Delta Kappan
  \textbf{47}(5),  238--243 (1966)

\bibitem{pane2014effectiveness}
Pane, J.F., Griffin, B.A., McCaffrey, D.F., Karam, R.: Effectiveness of
  cognitive tutor algebra i at scale. Educational Evaluation and Policy
  Analysis  \textbf{36}(2),  127--144 (2014)

\bibitem{gpt2}
Radford, A., Wu, J., Child, R., Luan, D., Amodei, D., Sutskever, I., et~al.:
  Language models are unsupervised multitask learners. OpenAI blog
  \textbf{1}(8), ~9 (2019)

\bibitem{CogTutor}
Ritter, S., Anderson, J.R., Koedinger, K.R., Corbett, A.: Cognitive tutor:
  Applied research in mathematics education. Psychonomic bulletin \& review
  \textbf{14}(2),  249--255 (2007)

\bibitem{roschelle2016online}
Roschelle, J., Feng, M., Murphy, R.F., Mason, C.A.: Online mathematics homework
  increases student achievement. AERA open  \textbf{2}(4),  2332858416673968
  (2016)

\bibitem{mathgpt}
Scarlatos, A., Lan, A.: Tree-based representation and generation of natural and
  mathematical language. In: Association for Computational Linguistics (ACL)
  (2023, preprint: \url{https://arxivorg/abs/230207974})

\bibitem{mathbert}
Shen, J.T., Yamashita, M., Prihar, E., Heffernan, N.T., Wu, X., Lee, D.:
  Mathbert: {A} pre-trained language model for general {NLP} tasks in
  mathematics education. CoRR  \textbf{abs/2106.07340} (2021),
  \url{https://arxiv.org/abs/2106.07340}

\bibitem{irtasag}
Uto, M., Uchida, Y.: Automated short-answer grading using deep neural networks
  and item response theory. In: International Conference on Artificial
  Intelligence in Education. pp. 334--339 (2020)

\bibitem{wang2019meta}
Wang, Z., Lan, A., Waters, A., Grimaldi, P., Baraniuk, R.: A meta-learning
  augmented bidirectional transformer model for automatic short answer grading.
  In: Proc. 12th Int. Conf. Educ. Data Mining (EDM). pp.~1--4 (2019)

\bibitem{treeEmbed}
Wang, Z., Lan, A.S., Baraniuk, R.G.: Mathematical formula representation via
  tree embeddings. In: iTextbooks@ AIED. pp. 121--133 (2021)

\bibitem{huggingFace}
Wolf, T., Debut, L., Sanh, V., Chaumond, J., Delangue, C., Moi, A., Cistac, P.,
  Rault, T., Louf, R., Funtowicz, M., Davison, J., Shleifer, S., von Platen,
  P., Ma, C., Jernite, Y., Plu, J., Xu, C., Scao, T.L., Gugger, S., Drame, M.,
  Lhoest, Q., Rush, A.M.: Huggingface's transformers: State-of-the-art natural
  language processing (2019). \doi{10.48550/ARXIV.1910.03771},
  \url{https://arxiv.org/abs/1910.03771}

\bibitem{edm22}
Zhang, M., Baral, S., Heffernan, N., Lan, A.: Automatic short math answer
  grading via in-context meta-learning. arXiv preprint arXiv:2205.15219  (2022)

\bibitem{MathOp}
Zhang, M., Wang, Z., Baraniuk, R.G., Lan, A.S.: Math operation embeddings for
  open-ended solution analysis and feedback. CoRR  \textbf{abs/2104.12047}
  (2021), \url{https://arxiv.org/abs/2104.12047}

\end{thebibliography}

\end{document}